\pdfoutput=1

\documentclass[11pt]{article}

\usepackage[]{acl}

\usepackage{times,xspace}
\usepackage{latexsym}

\usepackage{helvet} 
\usepackage{courier}  
\usepackage{graphicx} 
\usepackage{natbib}  
\usepackage{caption} 

\usepackage{graphicx}
\usepackage{booktabs}
\usepackage{subfigure}
\usepackage{algorithm}
\usepackage{algorithmic}
\usepackage{color}

\usepackage{microtype}
\usepackage{amsmath, amssymb, amscd, amsfonts}
\usepackage{IEEEtrantools}
\usepackage{soul}
\usepackage{url}
\usepackage{algorithm}
\usepackage{algorithmic}
\usepackage{amssymb}
\usepackage{amsfonts} 
\usepackage{multirow}
\usepackage{boldline}
\usepackage{hhline}
\usepackage{xcolor}

\usepackage{makecell}
\usepackage{mathrsfs}
\usepackage[normalem]{ulem}
\useunder{\uline}{\ul}{}
\usepackage[inline]{enumitem}
\usepackage{enumitem}

\usepackage[T1]{fontenc}

\usepackage[utf8]{inputenc}

\usepackage{microtype}

\title{Causal Intervention Improves Implicit Sentiment Analysis}

\author{Siyin Wang$^1$, {\bf Jie Zhou$^{1}$\thanks{\quad Jie Zhou is the corresponding authors of this paper.}} \and {\bf Changzhi Sun$^2$} \and {\bf Junjie Ye$^1$} \and \\ {\bf Tao Gui$^1$} \and {\bf Qi Zhang$^1$} \and {\bf Xuanjing Huang$^1$} \\
        $^1$School of Computer Science, Fudan Univerisity, Shanghai, China \quad\quad  $^2$ByteDance AI Lab \\
        \{siyinwang20, jie\_zhou, jjye19, tgui, qz, xjhuang\}@fudan.edu.cn, sunchangzhi@bytedance.com}

\newcommand{\mymodel}{\texttt{ISAIV}\xspace}

\begin{document}
\maketitle

\begin{abstract}
Despite having achieved great success for sentiment analysis, existing neural models struggle with implicit sentiment analysis.
This may be due to the fact that they may latch onto spurious correlations (“shortcuts”, e.g., focusing only on explicit sentiment words),
resulting in undermining the effectiveness and robustness of the learned model.
In this work,
we propose a causal intervention model for Implicit Sentiment Analysis using Instrumental Variable (\mymodel).
We first review sentiment analysis from a causal perspective and analyze the confounders existing in this task.
Then, we introduce an instrumental variable to eliminate the confounding causal effects, thus extracting the pure causal effect between sentence and sentiment.
We compare the proposed \mymodel  model with several strong baselines on both the general implicit sentiment analysis and aspect-based implicit sentiment analysis tasks.
The results indicate the great advantages of our model and the efficacy of implicit sentiment reasoning.
\end{abstract}

\section{Introduction}
\label{sec:intro}

The remarkable success that the field of sentiment analysis has achieved in the past few years has been derived from the use of increasingly high-capacity neural models to extract correlations from data \cite{MatthewEPeters2018DeepCW,JacobDevlin2018BERTPO,YinhanLiu2019RoBERTaAR}.
Although having reached state-of-the-art results, correlational predictive models can be untrustworthy
~\cite{RiccardoGuidotti2018ASO}: they may latch onto spurious correlations (“shortcuts”), leading to poor generalization.

\begin{figure}[t!]
\centering
\includegraphics[scale=0.5]{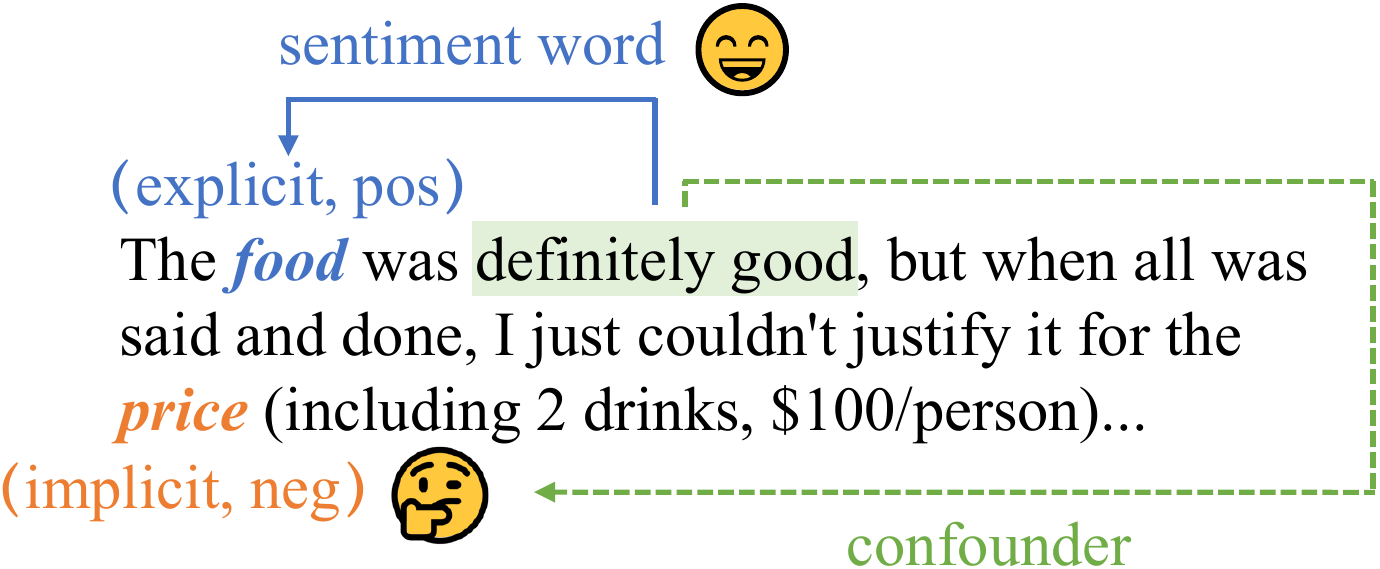}
\caption{An examples of confounding factors in implicit sentiment analysis for ABSA.}
\label{fig:example}
\end{figure}

One shortcut might be the explicit sentiment word which is 
a powerful feature cue.
Unfortunately, such a shortcut severely harms the generalization
and the robustness of the learned models in implicit sentiment analysis (ISA), where there are no explicit sentiment words about the topic or aspect \cite{russo2015semeval}.
Figure \ref{fig:example} gives a sample of aspect-based sentiment analysis (ABSA) \cite{zhou2020sk,zhou2020position}, which aims to predict the sentiments of the aspects in the sentence. The aspect ``food'' has explicit sentiment words ``definitely good'', but aspect ``price'' does not.
If the model thoughtlessly relies on shortcuts to sentiment words, it may make an incorrect sentiment prediction about the aspect ``price''.
In fact, there are many other kinds of shortcuts that models may learn,
for example, the rhetorical question mood expressed by the users ~\cite{ranganath2018understanding} and the co-occurrence of neutral words and sentiment polarities~\cite{wang2020identifying}.

The above shortcomings can potentially be addressed by the causal perspective: knowledge of causal relationships between observations and labels can be used to formalize spurious correlations and alleviate the predictor's dependence on them ~\cite{buhlmann2020invariance,veitch2021counterfactual,feder2021causal}.
Motivated by a causal perspective, we incorporate domain knowledge of the causal structure of the data into the learning objective.
Specifically, causal intervention is used to curb dependence on shortcuts (e.g., ``good $\rightarrow$ positive'') and improve the ability to reason \textit{causal effect} between sentence and sentiment. 

In this paper, we rethink the ISA task from a causal perspective and unitize the casual intervention on deep learning. 
We argue that the causal effect obtained by reasoning directly from the sentence ($X$) to the sentiment ($Y$) without relying on other extra prior stereotypical lexical impressions is closer to the original semantic analysis. Our work aims at eliminating the confounding causal effects of $C \rightarrow Y$ and thus extracting the pure causal effect between sentence and sentiment. 
Inspired by the instrumental variable in causality, we propose a causal intervention model for ISA using instrumental variable (\texttt{ISAIV}). Different from the other work with causal intervention like back-door adjustment \cite{landeiro2016robust}, other variables like confounders are not required to be observed. \texttt{ISAIV} contains two-stage learning: (1) In the first stage, we model the relationship between the instrumental variable and sentence; (2) In the second stage, we dismiss the spurious correlation between confounders and sentiment by means of the relationship obtained from the first stage.

To evaluate the effectiveness of our \texttt{ISAIV}, we conduct a series of experiments on both the general implicit sentiment analysis and aspect-based implicit sentiment analysis.
In particular, we compare our model with several the mainstream baselines and the results show the great advantages of our model on ISA.
We also validate the robustness of the model by adversarial attack and case studies, which proves that our model successfully dismisses the spurious correlation caused by sentiment words and extracts the pure causal effect. 

The main contributions are summarized as follows.
 \begin{itemize}[leftmargin=*, align=left]
     \item We rethink the implicit sentiment from a causal perspective and proposed a casual intervention model for implicit sentiment analysis (\texttt{ISAIV}).
    
     \item To remove the spurious causes of confounders, we incorporate instrumental variable into neural network to enhance its causal reasoning ability.
    
     \item We conduct experiments on diverse datasets, including partially implicit and totally implicit sentiment, which shows our effectiveness and rationality to reason implicit sentiment.
    
\end{itemize}

\section{Preliminaries}
\label{sec:prelim}

\subsection{Structural Causal Model and Causal Effect}
\label{sect:Structural Causal Model and Causal Effect}
\begin{figure}[t]
    \centering 
    \subfigure[Causal Graph]{
        \label{Fig.sub.1}
        \includegraphics[width=2.4cm,height = 2.8cm]{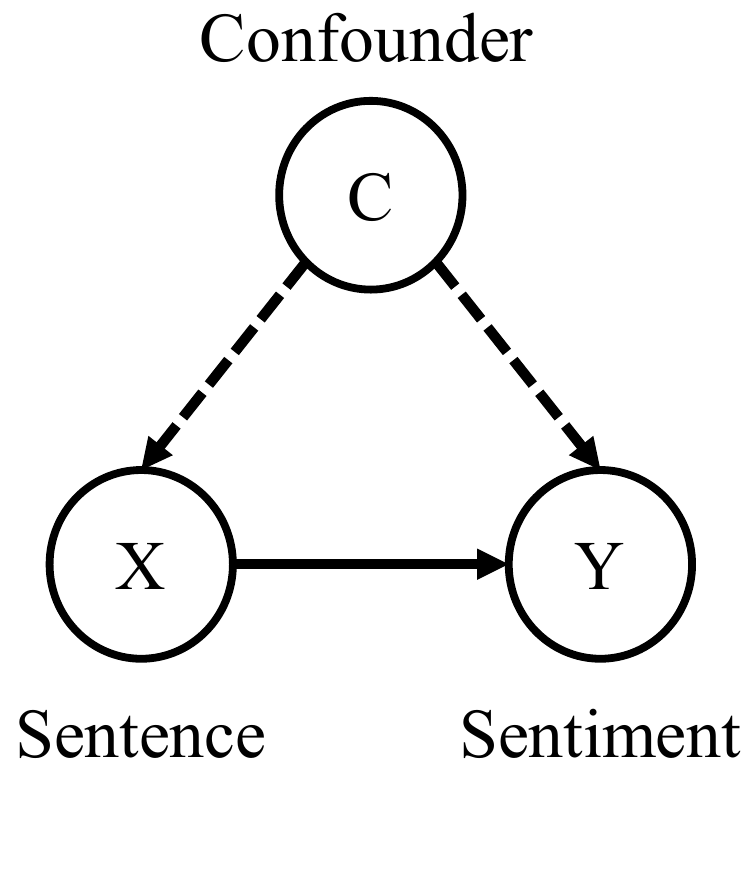}}
    \subfigure[Instrument Variable for Causal Intervention]{
        \label{Fig.sub.2}
        \includegraphics[width=4.2cm,height = 2.8cm]{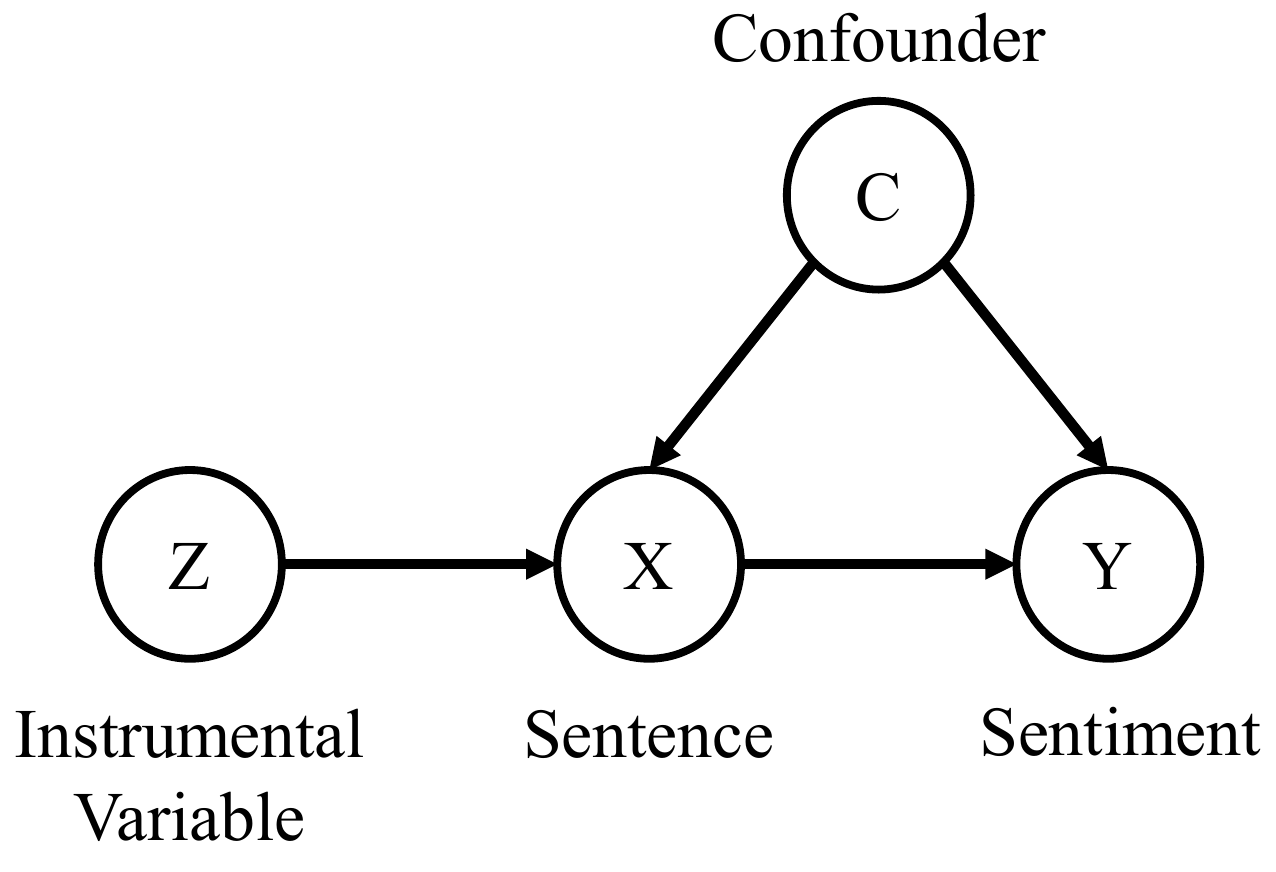}}
    \caption{Causal Graph}
    \label{fig:casual graph}
    \vspace{-4mm}
\end{figure}

In our paper, Structural Causal Model (SCM) \cite{glymour2016causal} is represented as a directed acyclic graphs (DAGs) $G = \left\{V, E\right\}$ to reflect causal relationships, where $V$ denotes the set of observational variables and $E$ denotes the direct causal effect (Figure \ref{Fig.sub.1}).
$X$ is a direct cause of $Y$ when variable $Y$ is the child of $X$. 

Variable $X$ and $Y$ is called treatment variable and outcome variable respectively when observing the causal relationship between them.
The other variables we do not consider their causal relationship are called  error terms ($\varepsilon$), also known as exogenous variables.
Significantly, total effect between $X$ and $Y$, denoted as $P(Y \mid X)$, is conceptually different from causal effect of $X \rightarrow Y$, denoted as $P(Y \mid do(X=x))$ because the causal effect only involves the direct path from $X$ to $Y$, while the total effect involves all paths between $X$ and $Y$. 
Based on the ideal hypothesis that none of the error terms will involve in the path between $X$ and $Y$, people usually treat the total effect and the causal effect as the same. But the actual fact is that a part of error terms (we call it confounder ($C$)) serves as a common cause of the treatment and outcome, denoted as $X \leftarrow C \rightarrow Y$. Consequently, the total effect is practically different from the causal effect, i.e. $P(Y \mid do(X=x)) \neq P(Y \mid X)$ and treatment-outcome relationship may well be obscured by the spurious correlation between $C$ and $Y$ generated by confounder \cite{pearl_2009,MA2020causalif}. 







\subsection{Instrument Variable for Causal Intervention}
\label{sect:Instrument Variable for Causal Intervention}
To recover the gap between total effect $P(Y \mid X)$ and casual effect $P(Y \mid do(X=x))$ and derive pure causal effect, we must ``adjust'' for potential confounder ($C$) \cite{pearl1995causal}.
Fortunately, applying causal intervention can extract the pure causality from the correlation and therefore overcome the problem of confounding bias. 
There are four key interventions: randomized controlled trial, backdoor adjustment, front-door adjustment, and instrumental variable estimation.
Randomized controlled trials are simply not practicable for natural language, and both the front-door and back-door adjustment require additional observable variables. 
However, the confounders (e.g., rhetoric confounding word, such as rhetorical questions and sarcasm) are too polymorphic to be observed exhaustively for implicit sentiment analysis.
We adopt the instrumental variable to dismiss the spurious correlations instead of directly observing confounders (Figure \ref{Fig.sub.2}).

Before the intervention, we should find a suitable instrumental variable ($Z$) that qualifies well the requirements as follows:\cite{brito2012generalized}

1. $Z$ is independent of all error terms $\varepsilon$ that have an influence on $Y$ which is not mediated by $X$, $Cov(Z, \varepsilon) = 0$.

2. $Z$ is not independent of $X$, $Cov(Z,X) \neq 0$.

The intuition behind this definition is that all correlation between $Z$ and $Y$ requires $X$ to act as an intermediary.


Generally, instrument variable estimation contains two stages \cite{angrist2008mostly}. In the first stage, the coefficient $\alpha$ is obtained by regression estimation of $X$ and $Z$, denoted as ${Cov(Z,X)}$. In the second stage, substitute $X$ with the expressions including $Z$ obtained in the first stage into the expression of $Y$ and then regress $Y$ on $Z$ 
, denoted as ${Cov(Z,Y)}$. 
There is no confounding bias between $Y$ and $Z$ owing to the restriction in the definition of $Z$, i.e. ${Cov(Z,\varepsilon)} = 0$ . 
A simple linear model for IV estimation consists of 2 equations:
\begin{equation}
\begin{aligned}
 X = \alpha Z + \varepsilon_X ; Y = \omega X + \varepsilon_Y
\end{aligned}
\end{equation}
where $Y$ is the outcome variable (e.g., sentiment), $X$ is the treatment variable (e.g, sentence), $Z$ is the instrumental variable (e.g., stochastic perturbation), and $\varepsilon_X$ and $\varepsilon_Y$ are error terms including but not limited to confounders(C). 
Under the conditions above, it can be proved that the equation presents an asymptotically unbiased estimate of the effect of $X$ on $Y$ \cite{angrist1996identification}.
\begin{equation}
    {\omega_{IV}} = \frac{\frac{1}{n}\sum_{i=1}^{n}(z_i-\overline{z})(y_i-\overline{y})}
    {\frac{1}{n}\sum_{i=1}^{n}(z_i-\overline{z})(x_i-\overline{x})}
    =\frac{Cov(Z,Y)}{Cov(Z,X)}
\label{equ:beta_iv}
\end{equation}


\begin{figure*}[t!]
\vspace{-6mm}
\centering
\includegraphics[scale=0.5]{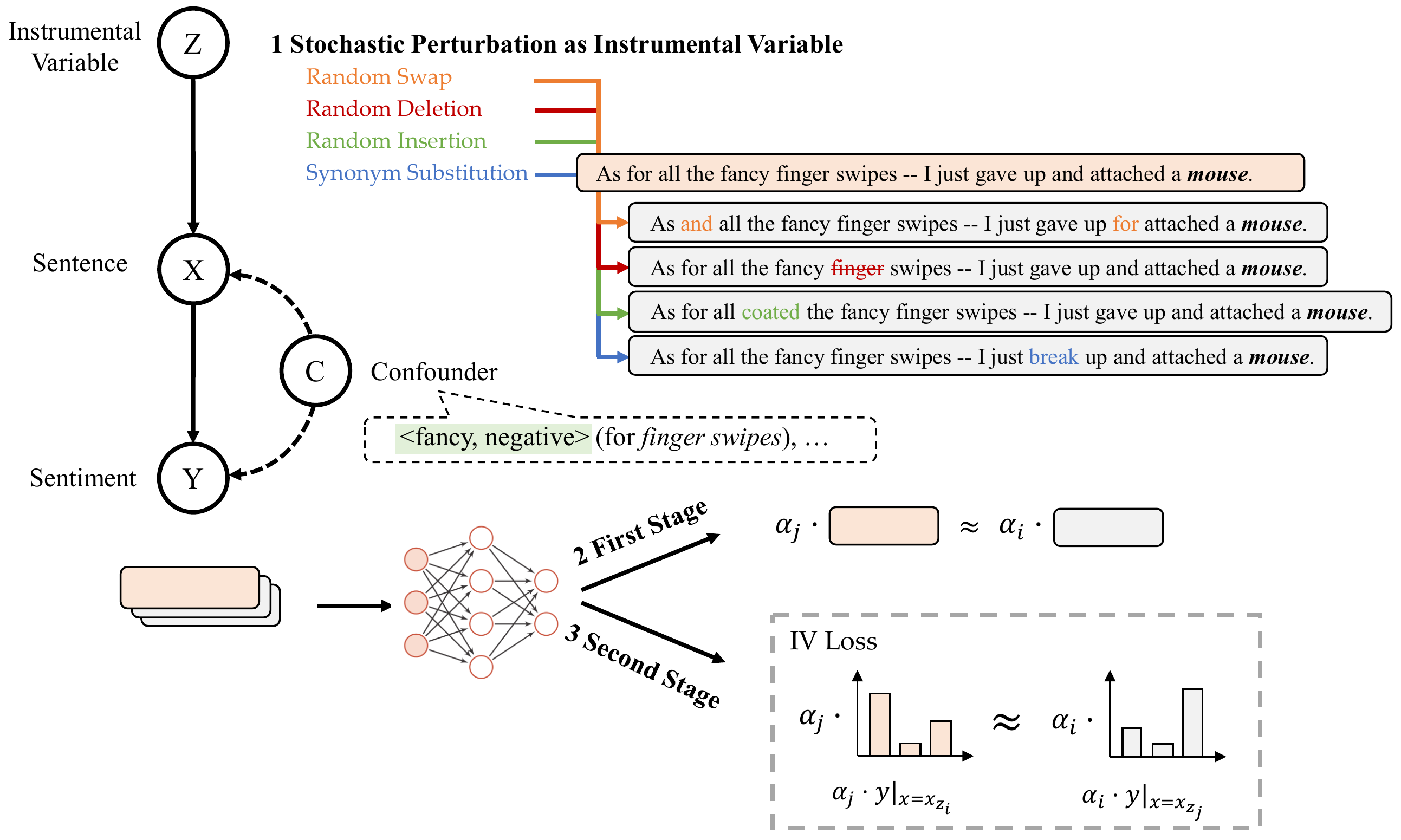}
\vspace{-1mm}
\caption{The framework of our \texttt{ISAIV}.}
\label{fig:framework}
\vspace{-3mm}
\end{figure*}


\section{Our Approach}
\label{sect:isaiv}
In this section, we introduce our \texttt{ISAIV} model for implicit sentiment analysis (Figure \ref{fig:framework}).
We first rethink the ISA from a causal perspective (Section \ref{sect:Sentiment Analysis from Casual Perspective}).
Then, we adopt stochastic perturbation as instrumental variable (Section \ref{sect:Stochastic Perturbation as Instrumental Variable}) and estimate instrumental variable in two stages (Section \ref{sect:The First Stage of ISAIV} and \ref{sect:The Second Stage of ISAIV}).

\subsection{Sentiment Analysis from Casual Perspective}
\label{sect:Sentiment Analysis from Casual Perspective}
Given a sentence $X$, consisting of a sequence of tokens $(x_1, x_2, ..., x_n)$, our task aims to analyze the polarity $Y$. 
For aspect-based sentiment analysis task, we concatenate the sentence and the aspect as the input $X=(x_1, x_2, ..., x_n, [SEP], a_1, a_2, ..., a_m)$.
In the current method, a deep neural network is used as a classifier to predict the sentiment polarity label as output and the sentence as input (as Equation \ref{equ:classifier}).

\begin{equation}
\label{equ:classifier}
    y =h(x;w) = W_{xy} \cdot x + \varepsilon_y 
\end{equation}
where $\varepsilon_y$ denotes as the error terms including confounders ($c$) and other errors ($\hat{\varepsilon_y}$). 

The prediction above is based on the hypothesis that error terms will not involve in the causal path between $X$ and $Y$ and ignore the influence of error terms mostly.
Nevertheless, several research has found that text classification systems based on neural networks are biased towards learning frequent spurious correlations \cite{leino2018feature}. 
It urges us to focus on the longtime unheeded but unavoidable existence of confounder ($C$) in error terms, which results in the overlooked gap between \textit{total effect} and \textit{causal effect}, denoted as the path $X \leftarrow C \rightarrow Y$.
The Equation \ref{equ:classifier} can be updated in consideration of confounder (c) (Equation \ref{equ:confounder}).
\begin{equation}
\label{equ:confounder}
    y ={h}(x;w)\\
    = W_{xy} \cdot x  + W_{cy} \cdot c + \hat{\varepsilon_y}
\end{equation}
where $c$ and $\hat{\varepsilon_y}$ denotes the confounder and other error terms respectively, $W_{cy}$ denotes the causal effect of $C \rightarrow Y$.

In previous studies, gender \cite{DBLP:conf/emnlp/FieldT20}, age, and address \cite{landeiro2016robust} were found to be confounders in text classification. 
As for ISA, we focus rather on the naturally existing confounder within the text, i.e., sentiment words. 
Sentiment words affect the form of the text as a component of the text (i.e., the writer's word choice determines the form of expression) and also affect sentiment expression \cite{xing2020tasty}. The existence of sentiment words as confounder makes it difficult to distinguish the pure causal effect of $X \rightarrow Y$ and the prediction indiscriminately depends on the spurious correlation between sentiment words and sentiment will fail in ISA. 
The main forms are as follows:

$\bullet$ \textbf{Inter-aspect Confounding Word} Explicit sentiment words of other aspects with opposite sentiment in the sentence confounds the prediction effect of the current aspect.
    
$\bullet$ \textbf{Inter-clause Confounding Word} In an adversative compound sentence, the other clauses with opposite semantics confound the prediction effect of the current clause. 
    
$\bullet$ \textbf{Rhetoric Confounding Word} 
Sentiment words conveying the opposite of the norm in rhetorical devices such as rhetorical questions and sarcasm confound the prediction effect.
    
$\bullet$ \textbf{Dynamic Neural Confounding Word} Neutral words show dynamic sentiment polarity in different contexts, but the model trained by biased data only learns the spurious correlation of one polarity.
  
We also provide a detailed analysis of the confounder in case studies (Section \ref{sec:case studies}).

Inspired by causal intervention with instrumental variable (Section \ref{sect:Instrument Variable for Causal Intervention}), we adopt two-stage instrument variable estimation for ISA to achieve the goal that distinguishes the pure \textit{causal effect} of $X\rightarrow Y$ without any spurious correlations, denoted as $P(Y\mid do(X=x))$.
\begin{table*}[t!]
\centering
\small
\caption{The statistics information of the datasets. IS means the percent of samples that are implicit sentiment.}
\vspace{-1mm}
\setlength{\tabcolsep}{3.5mm}{\begin{tabular}{lcccccccc}
\hlineB{4}
\multirow{2}{*}{Dataset} & \multicolumn{2}{c}{Postive} & \multicolumn{2}{c}{Neural} & \multicolumn{2}{c}{Negative} & \multicolumn{2}{c}{IS (\%)} \\
                         & Train         & Test        & Train        & Test        & Train         & Test         & Train               & Test                \\ \hline
Restaurant               & 2164          & 728         & 805          & 196         & 633           & 196          & 28.59               & 23.84               \\
Laptop                   & 987           & 341         & 866          & 128         & 460           & 169          & 30.87               & 27.27               \\
CLIPEval                 & 435           & 144         & 205          & 72          & 640           & 155          & 100.00              & 100.00     \\
\hlineB{4}
\end{tabular}}
\vspace{-1mm}
\end{table*}


\subsection{Stochastic Perturbation as Instrumental Variable}
\label{sect:Stochastic Perturbation as Instrumental Variable}
For text, the two restrictions of instrumental variable could be translated into two basic opinions: (1) instrumental variable $Z$ will not influence the sentiment polarity via any other casual path except through sentence $X$;
(2) instrumental variable $Z$ will influence the format of sentence $X$. 
Intuitively, we choose the stochastic perturbation as the instrumental variable of ISA. 
Inspired by the work of data augmentation \cite{LiuGuohang2020EasyDA}, we choose random swap, random deletion, random insertion, and synonym substitution as stochastic perturbation: 
A) \textbf{Random Swap}: Swap word randomly; B) \textbf{Random Deletion}: Delete word randomly with probability $p$; C) \textbf{Random insertion}: Insert word randomly by word embeddings similarity; D) \textbf{Synonym Substitution}: Substitute word by WordNet's synonym.
It fortunately meets the requirements of instrumental variable well: 
(1) stochastic perturbation obviously has no independent effect on sentiment, except through augmentation sentences, i.e. $ Cov(Z,\varepsilon) = 0$; 
(2) stochastic perturbation above will definitely change the sentence into another form, i.e. $Cov(Z,X) \neq 0$.

\subsection{The First Stage of ISAIV}
\label{sect:The First Stage of ISAIV}
Following the traditional pattern of instrumental variable estimation (Section \ref{sect:Instrument Variable for Causal Intervention}), the first stage of \texttt{ISAIV} is to establish the causal relationship between stochastic perturbation ($Z$) and sentence ($X$), i.e. $Z \rightarrow X$. 
We use two open source tools\footnote{\href{https://github.com/jasonwei20/eda_nlp}{https://github.com/jasonwei20/eda\_nlp}\\\href{https://github.com/makcedward/nlpaug}{https://github.com/makcedward/nlpaug}} to generate augmentation samples $x_z$ from the original sample $x$ and the formal expression is as follows.
\begin{equation}
\nonumber
    x_z = f(x,z)
\end{equation}
where $f(\cdot)$ denotes the different stochastic perturbation on the original sentence. 
For a specific stochastic perturbation $z_i$, we have $x_{z_i} = f(x,z_i) \approx \alpha_{i} \cdot x$. 
To obtain the accurate value of $\alpha$ which can well represent the relationship between $x$ and $x_z$, a neural network was constructed based on the BERT and $\alpha$ was set as a self-learning parameter.
\begin{equation}
\nonumber
    \begin{aligned}
    \alpha_i = \mathop{\min}\limits_{\alpha}\sum_{x_{z_i} = f(x,z_i)} \parallel \mathcal{M}(x_{z_i}) - \alpha \cdot \mathcal{M}(x) \parallel
    \end{aligned}
\end{equation}
where $\mathcal{M}$ denotes a text encoder (e.g., BERT).

\subsection{The Second Stage of ISAIV}
\label{sect:The Second Stage of ISAIV}
Substituting the relation above between original sample $x$ and augmentation sample $x_{z_i}$ into Equation \ref{equ:confounder}, we will get the $y \mid_{x=x_{z_i}}$ with different proportionality coefficient $\alpha$ obtained from the first stage. 
\begin{equation}
    \begin{aligned}
        y \mid_{x=x_{z_i}} &= W_{xy} \cdot x_{z_i}  + W_{cy} \cdot c + \hat{\varepsilon_y}\\
&= \alpha_{i} \cdot W_{xy} \cdot x + W_{cy} \cdot c+ \hat{\varepsilon_y}\\
&= \alpha_{i} \cdot y \mid_{do(x=x)}+ W_{cy} \cdot c+ \hat{\varepsilon_y}
    \end{aligned}
\label{equ:stage2y}
\end{equation}
where $y \mid_{do(x=x)}$ denotes the prediction only along the path $X \rightarrow Y$.

As the neural network is not totally linear, we slightly generalize the usage of two stages in instrumental variable.
We set the dismission of influence of the confounder as a regularization function instead of directly calculating the effect between $X$ and $Y$ as a single value (Equation \ref{equ:beta_iv}), which is obviously more fit for deep learning method.
\begin{equation}
\nonumber
    \begin{aligned}
    \mathcal{L}_{IV} &= \sum_{i \neq j} \parallel \alpha_j \cdot 
y\mid_{x=x_{z_i}} - \alpha_i \cdot 
y\mid_{x=x_{z_j}} \parallel \\
    \end{aligned}
\end{equation}

The reason we just model the prediction $y\mid_{x=x_{z_i}}$ and unitize the regularization loss on it is that the essence of the $\mathcal{L}_{IV}$ is to force the model to suppress the confounding effect caused by sentiment words. 
It can be easily proved by substituting the $y\mid_{x=x_{z_i}}$ with Equation \ref{equ:stage2y} obtained by two-stage learning. 
The benefit is obviously that we can suppress the confounding effect without directly observing the confounders (c). 
$$
\mathcal{L}_{IV}= \sum_{i \neq j} \parallel (\alpha_i-\alpha_j) \cdot (W_{cy} \cdot c+ \hat{\varepsilon_y}) \parallel
$$

In addition, the model should not go to the other extreme, i.e., ignore sentiment words entirely, which would be inconsistent with the normal process of natural language understanding. 
We set a hyperparameter $\beta$ to achieve balance and combine the causal regularization loss function $\mathcal{L}_{IV}$ with the conventional cross-entropy loss $\mathcal{L}_{CE}$ and the influence of the $\beta$ is analyzed in Section \ref{sect:influence of beta}.
\begin{equation}
\nonumber
    \mathcal{L}_{ALL} = \mathcal{L}_{CE} + \beta \mathcal{L}_{IV}
\end{equation}

\section{Experiment}
\label{sec:experiment}
\begin{table*}[t!]
\centering
\small
\caption{The main results for aspect-based sentiment analysis. For ESE and ISE, we provide the F1 score. We use the results reported in \cite{li-etal-2021-learning-implicit}. 
The baselines with $^\dag$ are our implementation.}
\vspace{-1mm}
\label{table:main results}
\setlength{\tabcolsep}{3mm}{\begin{tabular}{ll|cccc|cccc}
\hlineB{4}
\multicolumn{2}{l|}{}                                         & \multicolumn{4}{c|}{{Restaurant}}                                                                                     & \multicolumn{4}{c}{{Laptop}}                                                             \\
\multicolumn{2}{l|}{\multirow{-2}{*}{}}                       & {Acc}       & {F1}        & {ESE}       & ISE                              & {Acc} & {F1} & {ESE} & ISE                     \\ \hline
  & {ATAE-LSTM}    & 76.90  & 62.64  & 84.16  & 53.71 
                                        & 65.37  & 62.92  & 75.69  & 37.86\\
    & {IAN}          & 76.88  & 67.71  & 86.52  & 46.07  
                                        & 67.24  & 63.72  & 75.86  & 44.25\\
                                        
                       & {RAM}   & 80.23   & 70.80   & 85.11   & 55.81  
                                 & 74.49   & 71.35   & 75.86   & 44.25 \\
                       & {MGAN}    & 81.25   & 71.94   & 85.18   & 60.04  
                                   & 75.39   & 72.47   & 76.16   & 56.31 \\ \hline
\multirow{-4}{*}{NN}   & {TransCap}   & 79.55   & 71.41   & 86.52   & 59.93   
                                      & 73.87   & 70.10   & 77.16   & 60.34 \\
                       & {ASGCN}      & 80.77   & 72.02   & 84.29   & 62.91 
                                      & 75.55   & 71.05   & 75.46   & 57.77 \\
                       & {BiGCN}      & 81.97   & 73.48   & 87.19   & 59.05    
                                      & 74.59   & 71.84   & 79.53   & 62.64\\
                       & {CDT}        & 82.30   & 74.02   & 88.79   & 65.87                    & 77.19   & 72.99   & 77.53   & 68.90 \\ 
\multirow{-5}{*}{GNN}  & {RGAT}       & 83.30   & 76.08   & 89.45   & 61.05   
                                      & 77.42   & 73.76   & 80.17   & 65.52 \\\hline
    & {BERT-SPC}   & 83.57   & 77.16   & 89.21   & 65.54   
                                      & 78.22   & 73.45   & 81.47   & 69.54 \\
   & {CapsNet+BERT} & 85.09   & 77.75   & 91.68   & 64.04   
                                        & 78.21   & 73.34   & 82.33   & 67.24 \\
   & {BERT-PT}    & 84.95   & 76.96   & 92.15   & 64.79   
                                      & 78.07   & 75.08   & 81.47   & 71.27 \\
  & {BERT-ADA}   & \textbf{87.14}   & 80.05   & \textbf{94.14}   & 65.92   
                                      & 78.96   & 74.18   & \textbf{82.76}   & 70.11 \\
  & R-GAT+BERT   & 86.60   & 81.35   & 92.73   & 67.79   
                                      & 78.21   & 74.07   & 82.44   & 72.99 \\
  & TransEncAsp  & 77.10   & 57.92   & 86.97   & 48.96      & 65.83   & 59.53   & 74.31   & 43.20 \\
  & TransEncAsp+SCAPT  & 83.39  & 74.53  & 88.04  & 68.55 
                                      & 77.17   & 73.23   & 78.70   & 72.82 \\ \cline{2-10}
 & BERT-SPC$^\dag$ & 85.09 & 77.19 & 91.68 & 64.04 & 77.90 & 73.50 & 80.99 & 69.71 \\  
  & BERT-SPC$^\dag$ (Aug4) & 84.20   & 76.55   & 90.50   & 64.04    
                                      & 76.65   & 70.86   & 81.64   & 63.43 \\
     & BERT-SPC$^\dag$ (Aug8) & 80.98   & 67.77   & 90.39   & 50.94    
                                      & 75.71   & 71.62   & 77.97   & 69.71 \\
    \multirow{-11}{*}{BERT}        & BERT-SPC$^\dag$ (Aug16) & 77.59   & 67.44   & 85.35   & 52.81    
                                      & 74.61   & 69.92   & 77.97   & 65.71 \\        
                                    
\hline
\multirow{-1}{*}{Ours} & \texttt{ISAIV}   & {87.05}  & \textbf{81.40}  & 92.50  & \textbf{69.66}    &  \textbf{80.41}   & \textbf{77.25}   & 81.21                  & \textbf{78.29}  \\

\hlineB{4}
\end{tabular}}
\vspace{-1mm}
\end{table*}

\subsection{Datasets and Metrics}

\paragraph{Implicit Sentiment Analysis}
To show our model's better performance in understanding implicit sentiment, we evaluate the implicit polarity prediction on a total implicit dataset, CLIPEval from SemEval 2015 task 9 \cite{russo2015semeval}, which consists of self-reported entity reviews collected from psychological research with 1,280 sentences for the training and 371 for the test.

\paragraph{Aspect-based Implicit Sentiment Analysis}
 As our aim to dismiss the spurious correlation between explicit sentiment words and polarity, we also conducted experiments on both explicit and implicit datasets, Laptop and Restaurant review from SemEval 2014 task 4 \cite{MariaPontiki2014SemEval2014T4}. The segmentation of explicit sentiment (ESE) and implicit sentiment (ISE) is based on the work of \cite{li-etal-2021-learning-implicit} based on the annotation of opinion words \cite{ZhifangFan2019TargetorientedOW}.
 
We adopt two widely used metrics accuracy and macro-F1 to evaluate the performance of our model and the baselines.

\subsection{Baselines}
To investigate the effectiveness of our \texttt{ISAIV} model, we compare it with several typical baseline models for implicit sentiment analysis and aspect-based implicit sentiment analysis. 

\paragraph{Implicit Sentiment Analysis}
We select four popular baselines for implicit sentiment analysis, which are listed as follows.
SHELIFBK \cite{DBLP:conf/semeval/Dragoni15},
ATTLSTM \cite{DBLP:conf/iclr/LinFSYXZB17},
MTL \cite{DBLP:conf/ijcai/ZhengCQ18},
BERT-SPC \cite{xu-etal-2019-bert}.

\paragraph{Aspect-based Implicit Sentiment Analysis}
The commonly used baselines can be split into three parts, neural network, graph neural network, and BERT-based models, which are given as follows.

\textbf{Neural Network:}
ATAE-LSTM \cite{wang2016attention}, IAN \cite{DBLP:conf/ijcai/MaLZW17}, RAM \cite{DBLP:conf/emnlp/ChenSBY17}, MGAN \cite{DBLP:conf/emnlp/FanFZ18}.

\textbf{Graph Neural Network:}
TransCap \cite{DBLP:conf/acl/ChenQ19}, ASGCN \cite{DBLP:conf/emnlp/ZhangLS19}, BiGCN \cite{DBLP:conf/emnlp/ZhangQ20}, CDT \cite{DBLP:conf/emnlp/SunZMML19}, RGAT \cite{wang-etal-2020-relational}.

\textbf{BERT-based Models:}
BERT-SPC \cite{xu-etal-2019-bert}, CapsNet+BERT \cite{DBLP:conf/emnlp/JiangCXAY19}, BERT-ADA \cite{DBLP:conf/lrec/RietzlerSOE20}, R-GAT+BERT \cite{wang-etal-2020-relational}, TransEncAsp \cite{li-etal-2021-learning-implicit}, TransEncAsp+SCAPT \cite{li-etal-2021-learning-implicit}.

Moreover, to explore the influence of the augmentation sentences, we add them into the training dataset for our basic model (BERT-SPC).
For example, BERT-SPC (Aug4) means we add four augmentation samples for each example.
\subsection{Implementation Details}

We implement \texttt{ISAIV} with PyTorch based on Hugging Face Transformer \footnote{\href{https://huggingface.co/bert-base-uncased}{https://huggingface.co/bert-base-uncased}.}
and run them on the GPU(NVIDIA GTX 2080ti).
During training, we set the coefficient $\lambda$ of $\mathcal{L}_2$ regularization item is 0.01, $10^{-5}$ and dropout rate is 0.1. The learning rate is set as 2e-5 and the batch size is set as 16. Adam optimizer \cite{kingma2014adam} is used to update all the parameters.

\section{Experimental Analysis}

\begin{figure*}[t!]
\centering
\includegraphics[scale=0.5]{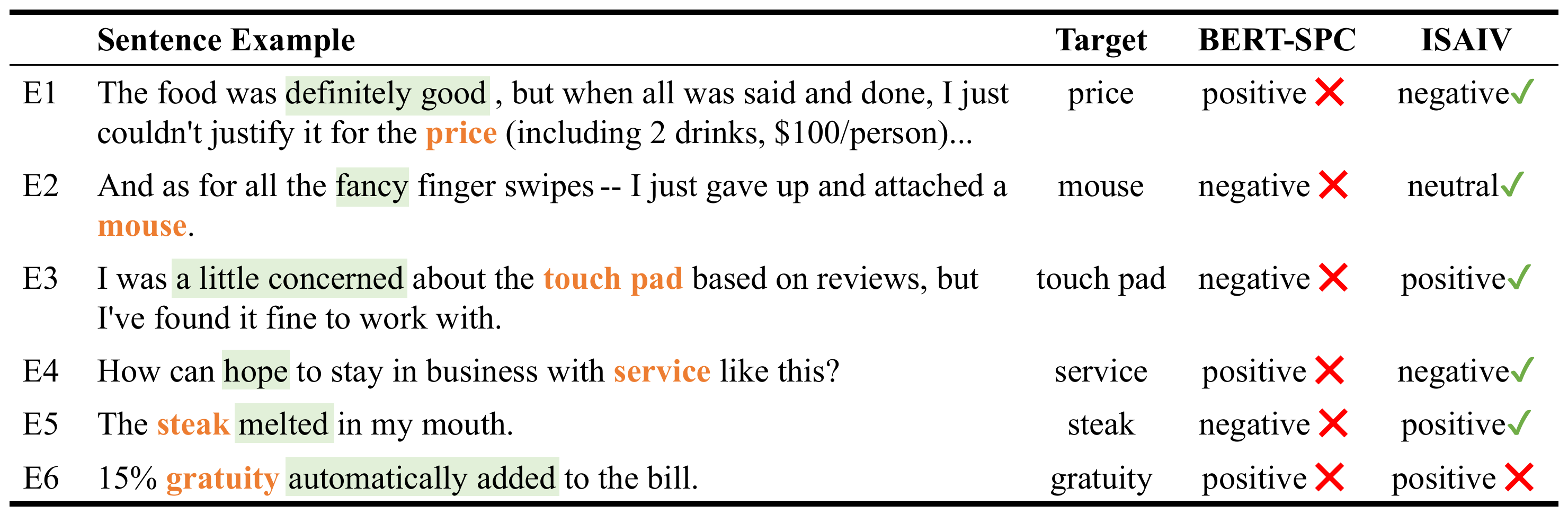}
\vspace{-1mm}
\caption{Some examples of case studies.}
\label{fig:case}
\vspace{-1mm}
\end{figure*}

\subsection{Main Results}
To evaluate the performance of our \texttt{ISAIV} model, we compare it with several mainstream baseline models for both the implicit sentiment analysis and aspect-based implicit sentiment analysis (Table \ref{table:main implicit} and Table \ref{table:main results}). 
We find the following observation from these tables.
\textbf{First}, our model outperforms all the baselines in most cases. 
Particularly, our model obtains the best F1 scores over all the three datasets of two tasks. 
\textbf{Second}, our \texttt{ISAIV} strategy significantly improves the performance of the baseline. 
\texttt{ISAIV} improves more than two points in terms of F1 over all the datasets compared with BERT-SPC, which is the base of our model.
\textbf{Third}, our model can improve the performance of implicit sentiment analysis effectively. 
For example, compared with the BERT-SPC$^{\dag}$, we obtain more than five points improvement on ISE over both Restaurant and Laptop. 
Also, we obtain the best results of implicit sentiment analysis over CLIPEval.
\textbf{Forth}, the model that regards the augmentation sentence as a data augmentation (e.g., BERT-SPC$^{\dag}$ (Aug4)) performs even worse than the one without augmentation as noise may exist.
This shows that our \texttt{ISAIV} algorithm can help learn the implicit sentiment reasoning behind the data.

\begin{table}[t!]
\centering
\small
\caption{The main results for implicit sentiment analysis. We use the results reported in \cite{xiang2021message}. The baselines with $^\dag$ are our implementation.}
\label{table:main implicit}
\setlength{\tabcolsep}{4.0mm}{\begin{tabular}{l|cc}
\hlineB{4}
\multirow{2}{*}{Method} & \multicolumn{2}{c}{CLIPEval}    \\
                        & Acc            & F1             \\ \hline
SHELLFBK                & 56.00          & 54.00          \\
ATTLSTM                 & 82.43          & 82.21          \\
MTL                     & 82.94          & 83.17          \\
BERT-SPC$^\dag$                & 87.06          & 84.74               \\ 
BERT-SPC$^\dag$ (Aug4)         & 85.71 & 83.56              \\ 
BERT-SPC$^\dag$ (Aug8)         & 86.52 & 85.30              \\ 
BERT-SPC$^\dag$ (Aug16)         & 85.44 & 84.54              \\ 
\hline
\texttt{ISAIV}          & \textbf{88.95} & \textbf{87.49} \\ 
\hlineB{4}
\end{tabular}}
\end{table}

\subsection{Case Studies}
\label{sec:case studies}
We present five samples in Figure \ref{fig:case} to explain the four main types of confounders (Section \ref{sect:Sentiment Analysis from Casual Perspective}), which shows the effectiveness and rationality of our model to reason implicit sentiment.
(1) \textbf{Inter-aspect Confounding Word.} 
In example E1, ``definitely good'' is the sentiment words of aspect \textit{food}, implying positive sentiment but confounds the prediction of aspect \textit{price}. 
In E2, the user expresses a negative sentiment towards aspect \textit{finger swipes} with opinion word ``fancy'', which confounds the prediction of aspect \textit{mouse}. 
(2) \textbf{Inter-clause Confounding Word.} In E3, the first and second clauses form an adversative relation, and the true meaning of the expression is that the \textit{mouse} works well, but the sentiment word ``a little concerned'' in the first clause confounds the prediction.
(3) \textbf{Rhetoric Confounding Word.} In E4, the customer used the rhetorical device of a rhetorical question to express that the restaurant's service was terrible, but the existence of the word ``hope'' confounds the prediction,
(4) \textbf{Dynamic Neural Confounding Word.} In E5, the word ``melted'' is absolutely a neutral word, but when we directly count the proportion of aspect-level sentiment polarity that co-occur with ``melted'', we surprisingly find that $83.33\%$ aspect polarity is negative, which well explains why the previous model predicts ``negative'' strangely. Due to the unbalanced distribution of training data, the model tends to tag the neural word with specific sentiment polarity and predict based on this spurious correlation learned superficially before.

\begin{table*}[t!]
\vspace{-6mm}
\centering
\small
\caption{The influence of augmentation sample number.}
\label{table:generated sample number}
\vspace{-1mm}
\setlength{\tabcolsep}{4mm}{\begin{tabular}{l|cccc|cccc}
\hlineB{4}
\multirow{2}{*}{\#Num} & \multicolumn{4}{c|}{Restaurant 0.6}                               & \multicolumn{4}{c}{Laptop 0.4}                                \\
                                       & Acc            & F1             & ESE            & ISE            & Acc            & F1             & ESE            & ISE                   \\ \hline
4                                      & 86.88          & 80.99          & 92.50          & 68.91          & \textbf{80.41} & \textbf{77.25} & \textbf{81.21} & \textbf{78.29} \\
8                                      & \textbf{87.05} & \textbf{81.40} & \textbf{92.50} & \textbf{69.66} & 78.68          & 75.23          & 79.70          & 76.00         \\
16                                     & 85.09          & 77.75          & 91.21          & 65.54          & 78.06          & 75.04          & 79.05          & 75.43         \\ \hlineB{4}
\end{tabular}}
\vspace{-1mm}
\end{table*}

\begin{figure}[t]
    \centering 
    \subfigure[Laptop]{
        \label{laptop}
        \includegraphics[width=3.5cm,height = 3.5cm]{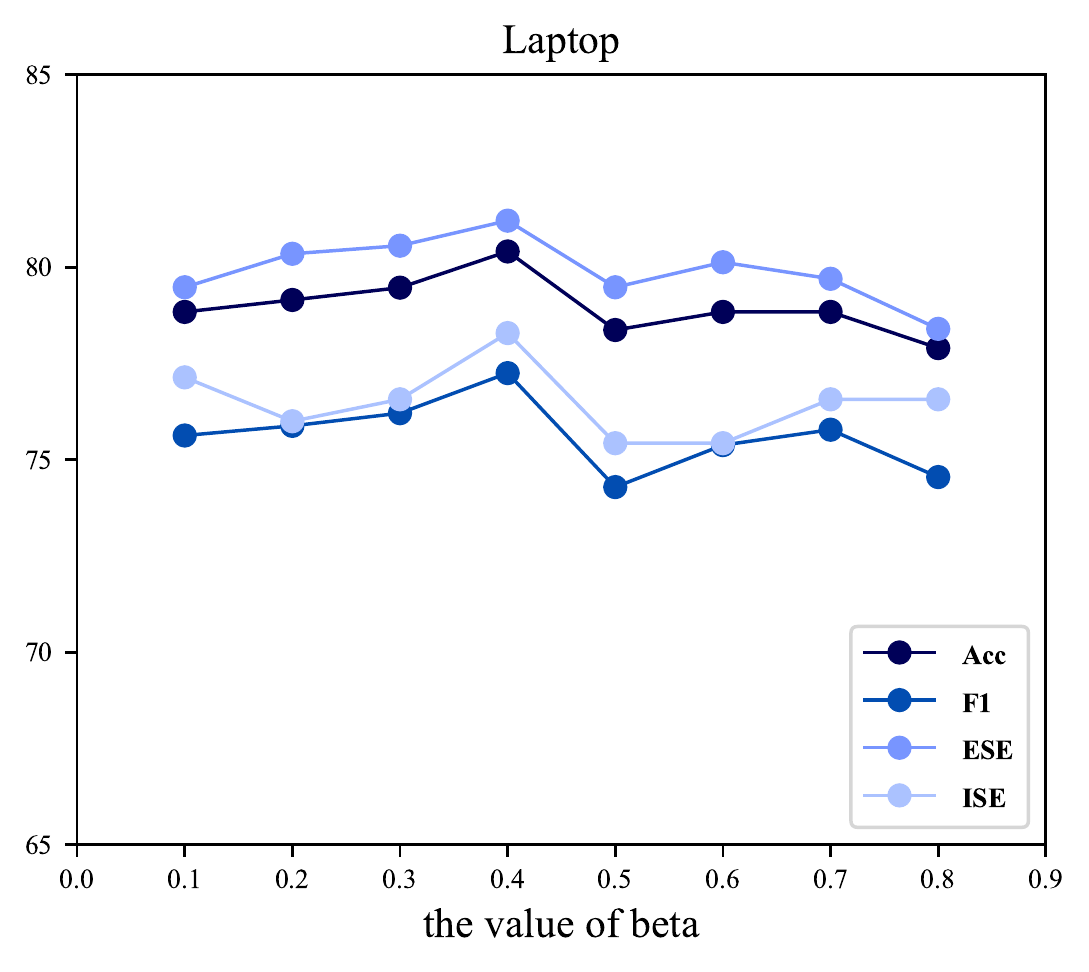}}
    \subfigure[CLIPEval]{
        \label{CLIPEval}
        \includegraphics[width=3.5cm,height = 3.5cm]{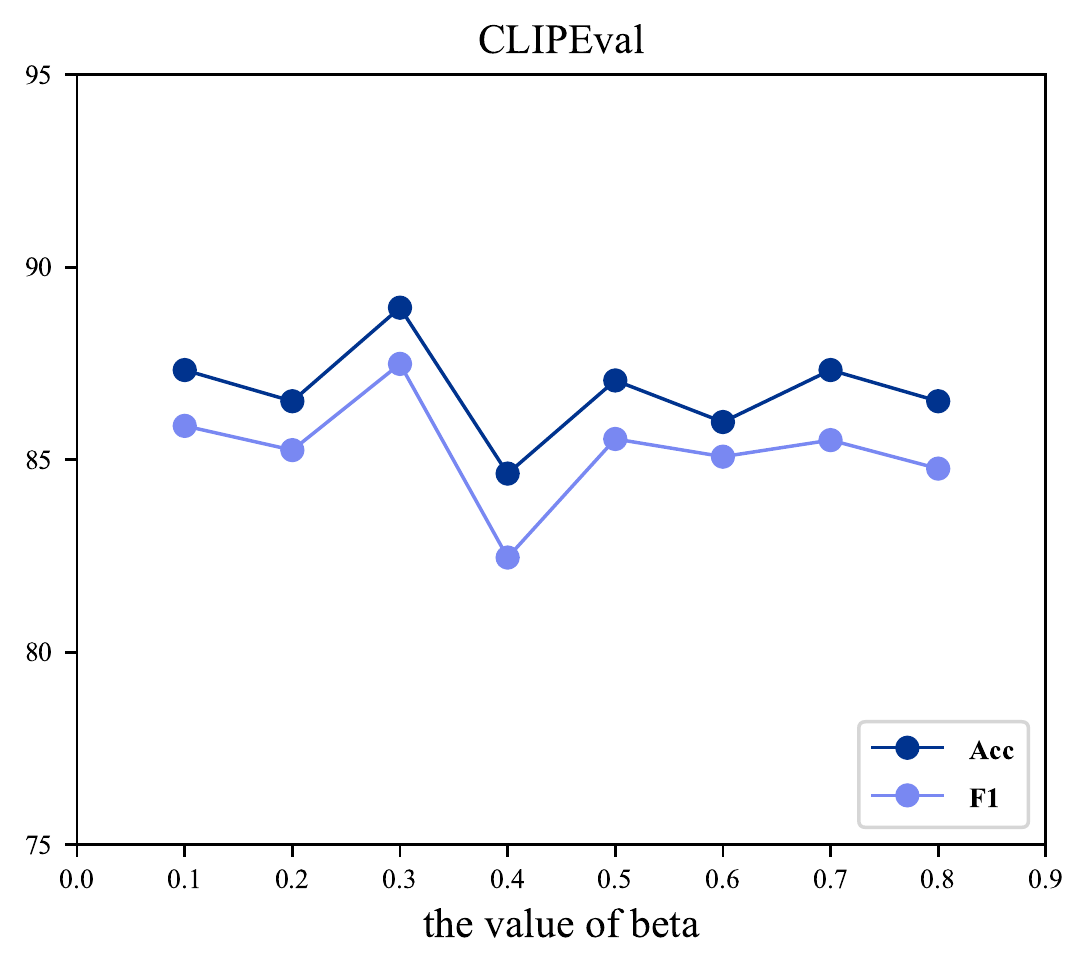}}
        \vspace{-2mm}
    \caption{The influence of $\beta$}
    \label{fig:influenceofbeta}
\end{figure}

\subsection{Further analysis}
\paragraph{Influence of Augmentation Sample Number.}
We explore the influence of augmentation sample number here (Table \ref{table:generated sample number}).
The influence of sample number on model performance depends on two conflicting factors: the degree of deviation from the original sentence and the chance to find more potential confounders. 
With the increase in sample number, the model has a greater chance of finding more potential confounders and adjusting for them. 
On the other hand, a larger generation samples number means that more samples deviating from the original sentence are involved in the learning procedure, and therefore the accuracy of prediction decreases. 
Over Restaurant, the two conflicting factors reach a better balance at 8; while on Laptop, the negative effect of semantic deviation outweighs the positive effect of correction for more confounders, and the best result is achieved at 4.

\paragraph{Influence of $\beta$.} 
Either emphasizing sentiment words only or completely ignoring them is not reasonable.
The purpose of our hyper-parameter beta is to strike a balance between these two terms (Figure \ref{fig:influenceofbeta}). 
In Laptop and CLIPEval, performance is best at 0.4 and 0.3 relatively, and both show a trend of high in the middle and low on both sides, indicating that our hypothesis is rational.

\label{sect:influence of beta}

\paragraph{Robustness.} 
We also analyze the robustness of our proposed \texttt{ISAIV} (Table \ref{table:robustness}).
We test our model on a robustness testing dataset, Revnon of TextFlint \cite{wang2021textflint}, which reverses the sentiment of the non-target aspects with originally the same sentiment as target. 
Our model outperforms the model BERT-SPC without causal intervention, which means \texttt{ISAIV} can also improve the robustness by learning the implicit sentiment reasoning.

\paragraph{Limitation.} 
We also analyze wrong samples and find the model may fail when encountering the expression with unusual knowledge. In Figure \ref{fig:case} E6, due to the lack of prior knowledge about ``gratuity", ``automatically added" is likely to be perceived as a good thing. 
Admittedly, our work mainly focuses on the reasoning ability and doesn't integrate external corpus and knowledge and therefore lacks abundant prior knowledge. The better combination of prior knowledge and causal inference is also an intriguing and worth exploring field.



\begin{table}[!t]
\small
\centering
\caption{The results of robustness.}
\label{table:robustness}
\setlength{\tabcolsep}{1.5mm}{\begin{tabular}{@{}l|cccc@{}}
\hlineB{4}
\multirow{2}{*}{Model} & \multicolumn{2}{c}{Restaurant (Trans.)} & \multicolumn{2}{c}{Laptop (Trans.)} \\
         & Acc   & F1    & Acc   & F1    \\ \hline
BERT-SPC & 57.04 & 44.43 & 51.05 & 41.01 \\
\texttt{ISAIV}     & 58.77 & 48.56 & 59.45 & 43.24 \\ \hlineB{4}
\end{tabular}}
\end{table}

\section{Related Work}
\label{sec:related}

\subsection{Implicit Sentiment Analysis}
Implicit sentiment analysis (ISA) task plays an important role in sentiment analysis field \cite{liu2012sentiment,zhou2019deep,zhou2020sentix}.
Early studies mainly trained machine learning models based on hand-crafted features or explicit characterization of implicit feature information. 
Some studies argued that seemingly neural words actually contain emotional content and then construct a lexicon~\cite{feng-etal-2013-connotation,castello-stede-2017-extracting}. 
Label propagation was used to judge the affective polarity of the words automatically \cite{ding2016acquiring,li2021iswr}.
Moreover, \citet{balahur2011detecting} proposed to build a commonsense knowledge base (EmotiNet) with the concept of affective value and the sentiment.

Recent efforts~\cite{he-etal-2018-effective,tang-etal-2020-dependency} used syntax information from dependency trees to enhance attention-based models. 
Using syntactic analysis tree and CNN, \citet{liao2019identification} analyzed fact-implied implicit sentiment by fusing multi-level semantic information, including sentiment target, sentence, and context semantic. 
A lot of works~\cite{DBLP:conf/emnlp/ZhangLS19,DBLP:conf/emnlp/SunZMML19,wang-etal-2020-relational} incorporated tree-structured syntactic information via graph neural networks to capture aspect-aware information in text.
Another method is to utilize external corpus and pre-trained knowledge to enhance semantic awareness of models~\cite{xu-etal-2019-bert,DBLP:conf/lrec/RietzlerSOE20,dai-etal-2021-syntax,li-etal-2021-learning-implicit,zhou2020sk}.

The existing methods mainly improve the ISA by integrating external corpus and knowledge.
However, the knowledge is always not complete which will influence the models' performance. In this paper, we solve it via causal intervention to learn the reasoning behind the sentiment classification.


\subsection{Causality for NLP}
Recently, some researchers are beginning to combine causality and NLP to create more robust and interpretable models~\cite{wood2018challenges,tang2021adversarial}.
Most of the papers integrated backdoor and counterfactual into NLP tasks.
Particularly, \citet{landeiro2016robust} applied the back-door adjustment to text classification by controlling the artificially predetermined confounding variable.
\citet{feng-etal-2021-empowering} introduced counterfactual reasoning into the model learning process by generating representative counterfactual samples and comparing the counterfactual and factual samples.
\citet{veitch2021counterfactual} utilized distinct regularization schema for distinct causal structure to induce counterfactual invariance.
\citet{niu2021counterfactual} utilized the counterfactual inference on VQA models by subtracting the language bias as direct language effect from the total causal effect.

Different from these studies, we explore the causal graph for ISA and incorporate it using the causal intervention.


\section{Conclusion}
\label{sec:conclusion}
In this paper, we proposed a causal intervention model for implicit sentiment analysis using instrument variable (\texttt{ISAIV}). Given that the current model indiscriminately focuses on the correlation between sentiment and sentiment words and consequently performs poorly in implicit sentiment analysis as the explicit sentiment words disappear, we rethink the implicit sentiment analysis from a causal perspective and analyze the four main forms of sentiment words as potential confounders. Inspired by the instrumental variable of causal intervention, we adopt stochastic perturbation as instrumental variable and construct a model with two-stage learning. Across three different datasets, including general implicit sentiment analysis and aspect-based sentiment analysis, our \texttt{ISAIV} shows great advantages in implicit sentiment.

\section*{Acknowledge}
The authors wish to thank the reviewers for their helpful comments and suggestions. This work was partially National Natural Science Foundation of China (No. 61976056, 62076069), Shanghai Municipal Science and Technology Major Project (No.2021SHZDZX0103).

\bibliography{main}
\bibliographystyle{acl_natbib}




\end{document}